\documentclass[conference]{IEEEtran}
\IEEEoverridecommandlockouts
\usepackage{amsmath,amssymb,amsfonts}
\usepackage{algorithmic}
\usepackage{graphicx}
\usepackage{textcomp}
\usepackage{xcolor}
\usepackage{bm}
\usepackage{hyperref}
    
\definecolor{ballblue}{rgb}{0.13, 0.67, 0.8}

\newcommand{\method}{V-ABC}
\newcommand{\xx}{\mathbf{x}}
\newcommand{\zz}{\mathbf{z}}
\newcommand{\XX}{\mathbf{X}}
\newcommand{\setting}{negative-unlabeled}
\newcommand{\KL}{\textrm{KL}}
\newcommand{\bphi}{\boldsymbol{\phi}}
\newcommand{\btheta}{\boldsymbol{\theta}}
\newcommand{\qzx}{q_{\bphi}(\mathbf{z}|\mathbf{x})}
\newcommand{\pxz}{p_{\btheta}(\mathbf{x}|\mathbf{z})}

\begin{document}

\title{Novel Applications for VAE-based Anomaly Detection Systems\\
}


\author{\IEEEauthorblockN{Luca Bergamin}
\IEEEauthorblockA{\textit{Department of Mathematics} \\
\textit{University of Padova}\\
Padova, Italy \\
\texttt{bergamin@math.unipd.it}}
\and
\IEEEauthorblockN{Tommaso Carraro}
\IEEEauthorblockA{
\textit{Fondazione Bruno Kessler}\\
Trento, Italy \\
\texttt{tcarraro@fbk.eu}}
\and
\IEEEauthorblockN{Mirko Polato}
\IEEEauthorblockA{\textit{Department of Computer Science} \\
\textit{University of Turin}\\
Turin, Italy \\
\texttt{mirko.polato@unito.it}}
\and
\IEEEauthorblockN{Fabio Aiolli}
\IEEEauthorblockA{\textit{Department of Mathematics} \\
\textit{University of Padova}\\
Padova, Italy \\
\texttt{aiolli@math.unipd.it}}
}

\maketitle

\begin{abstract}
	The recent rise in deep learning technologies fueled innovation and boosted scientific research. Their achievements enabled new research directions for deep generative modeling (DGM), an increasingly popular approach that can create novel and unseen data, starting from a given data set. As the technology shows promising applications, many ethical issues also arise. For example, their misuse can enable disinformation campaigns and powerful phishing attempts. Research also indicates different biases affect deep learning models, leading to social issues such as misrepresentation. In this work, we formulate a novel setting to deal with similar problems, showing that a repurposed anomaly detection system effectively generates novel data, avoiding generating specified unwanted data. We propose Variational Auto-encoding Binary Classifiers (V-ABC): a novel model that repurposes and extends the Auto-encoding Binary Classifier (ABC) anomaly detector, using the Variational Auto-encoder (VAE). We survey the limitations of existing approaches and explore many tools to show the model's inner workings in an interpretable way. This proposal has excellent potential for generative applications: models that rely on user-generated data could automatically filter out unwanted content, such as offensive language, obscene images, and misleading information.
\end{abstract}


\section{Introduction}

Deep Generative Models (DGMs) are on the rise \cite{goodfellow2014generative, kingma2014autoencoding}. Powered by deep learning, DGM had a remarkable impact, as these models nowadays can generate convincing people’s faces \cite{GANfaces,Maale2019BIVAAV} and news articles \cite{openai2019gpt2}. Novel applications are still researched, as they have been proven effective in many different tasks, such as model-based reinforcement learning and semi-supervised learning \cite{goodfellow2017nips}. 

However, this research community has brought up specific issues: having tools to control the generative capabilities of a DGM is crucial from an ethical and social standpoint. For example, state-of-the-art language models, despite having near human-like performance for specific tasks, have a troublesome behavior: they have been shown to favor misrepresentation, amplify social biases, favor stereotypes and even generate derogatory language \cite{gebru}. Trivial filtering techniques on training data have not been effective in dealing with the issue \cite{gebru}. 
Moreover, powerful language models can be abused. Their misuse can enable automated disinformation campaigns and powerful phishing attempts. Research organizations in the world opted for locking such models behind APIs out of fear of abuse \cite{openai2019gpt2}. We firmly believe that innovation cannot be driven by an \emph{opaque} system: we should be able to independently assess learning models to constantly improve them. Ideally, we should strive to create generative models that can be instructed to avoid generating specified inappropriate data, removing the capability to generate it altogether.

Anomaly detection systems based on Variational Autoencoders (VAEs) \cite{An2015VariationalAB,Xu2018UnsupervisedAD} have been a solid inspiration for this task. Although they are generally employed to perform classification, their generative formulation enables them to perform sampling. Moreover, their neural architecture can be easily adapted to different domains. VAE-based anomaly detection systems have been proved effective and resilient, showing to outclass PCA-based methods \cite{An2015VariationalAB}, random forests \cite{Xu2018UnsupervisedAD}, classic deep neural networks and one-class SVM \cite{yamanaka2019autoencoding}. 

Following these research directions, the main contributions of this paper are two-fold: (1) we propose a novel \textit{\setting{}} setting in which generative methods can be instructed to avoid the generation of unwanted data (e.g., outliers, malicious data); (2) we extend an autoencoder-based anomaly detector to a VAE inspired formulation to achieve this ambitious goal.
Understanding and promoting this under-explored setting is important for real-world applications that require a specific removal of sensitive topics from DGM, such as language modeling \cite{gebru} and image synthesis \cite{GANfaces}.

The remainder of the paper is structured as follows. In Section~\ref{sec:related} we present a brief survey about the applications of generative methods in settings similar to the proposed one. Section~\ref{sec:background} summarizes important notions and terminology which is used throughout the paper. Section~\ref{sec:method} presents our novel \setting{} setting and includes a detailed description of the proposed V-ABC model. Section~\ref{sec:experiments} presents a comparison between a standard VAE and V-ABC in our novel setting, using both toy and real-world data sets. Section~\ref{sec:results} discusses the results we have obtained and shows the ability of V-ABC in avoid generating unwanted data. Section~\ref{sec:conclusions} wraps up the paper and presents possible future directions.

\section{Related Work}\label{sec:related}

Anomaly detection systems, also known as \textit{outlier analysis}, have been researched for decades. These approaches have been successfully applied in many different scenarios, such as quality control, finance, cybersecurity, medical diagnosis, social monitoring, earth science, and data cleaning \cite{Aggarwal2013OutlierA}.

Variational autoencoders (VAEs) are more recent \cite{kingma2014autoencoding}, and many applications are still being discovered to this day. They have been successfully applied to anomaly detection tasks, in both supervised \cite{cern, Kawachi2018ComplementarySV} and unsupervised \cite{An2015VariationalAB,Xu2018UnsupervisedAD} settings.

A common solution for constrained sampling uses fully labeled data to produce samples with a specified class \cite{cvae}. This model is ineffective in the setting we propose, since the labeling process is noisy and it would aim to reproduce the same noisy distribution.

Considering negative constraints for generative models has been researched using statistical learning tools \cite{hanneke2018actively}, formulating the \textit{Generative Modeling with Negatives} task. Their contribution is mainly theoretical, while our work has a stronger applicative focus.

Alternative settings for generative models that learn from partially labeled data sets have been studied for the \textit{positive-unlabeled} case \cite{basile2017generative}, reporting a setting that requires generalizing the positive concept starting from a small positive set and a larger unlabeled data set for generative modeling (e.g., to perform density estimation).


\section{Background}\label{sec:background}
In this section, we provide some useful notation used throughout the paper and present the background knowledge that forms the basis of our method.

\subsection{Notation}
Bold notation is used to differentiate between vectors, e.g., $\mathbf{x} = [3.2, 2.1]$, and scalars, e.g., $\alpha = 5$. Probability distributions are denoted with lower case letters, e.g., $p(\cdot)$. Parametric distributions are denoted as $p_{\boldsymbol{\theta}}(\cdot)$, where $\boldsymbol{\theta}$ is a vector of parameters. With $\mathbf{z} \sim p(\mathbf{z})$, we indicate a sampling of some vector $\mathbf{z}$ from distribution $p(\mathbf{z})$. $p(\mathbf{x} | \mathbf{z})$ denotes the probability distribution of $\mathbf{x}$ conditioned by $\mathbf{z}$.
A training set $\mathbf{X}$ of $N$ examples is denoted as $\mathbf{X} = \{\mathbf{x}^{(i)}\}_{i=1}^N$, where $\mathbf{x}^{(i)}$ is the $i_{th}$ example of $\mathbf{X}$.

\subsection{Autoencoders}

Autoencoders (AEs) are unsupervised machine learning models based on neural networks. Their purpose is to learn an identity function that reconstructs the original input while compressing data in the process. These models are based on two neural networks, namely an encoder $f_{\boldsymbol{\phi}}$ and a decoder $g_{\boldsymbol{\theta}}$, parameterized by $\boldsymbol{\phi}$ and $\boldsymbol{\theta}$, respectively. The encoder learns a function which maps an high-dimensional input $\xx$ into a lower dimensional vector $\zz$, while the decoder learns how to reconstruct the original input $\xx$ from $\zz$. We refer to the reconstructed version of $\xx$ with $\tilde{\xx}$. So, given an input $\xx$, an autoencoder learns functions $f_{\boldsymbol{\phi}}$ and $g_{\boldsymbol{\theta}}$ such that $f_{\boldsymbol{\phi}}(g_{\boldsymbol{\theta}}(\xx)) = \tilde{\xx} \approx \xx$.

\subsection{Variational autoencoders}

Variational autoencoders are generative models based on variational inference, with an architecture similar to vanilla autoencoders. VAEs assume the input $\xx \in \XX$ is generated according to the following generative process: $\zz \sim p_{\boldsymbol{\theta}^*}(\zz)$ and $\xx \sim p_{\boldsymbol{\theta}^*}(\xx|\zz)$, with $\operatorname{dim}(\zz) \ll \operatorname{dim}(\xx)$. In other words, VAEs assume that the input vector $\xx$ is modeled as a function of an unobserved random vector $\zz$ of lower dimensionality. 

The objective of VAE is to estimate parameters $\boldsymbol{\theta}^*$ by maximizing the likelihood of the data (Maximum Likelihood Estimation, MLE), i.e., $\hat{\boldsymbol{\theta}}=\arg \max_{\boldsymbol{\theta} \in \boldsymbol{\Theta}}  \: p_{\boldsymbol{\theta}}(\xx)$.

Computing the MLE requires solving $ p_{\boldsymbol{\theta}}\left(\xx\right)=\int p_{\boldsymbol{\theta}}(\xx | \zz) p_{\boldsymbol{\theta}}(\zz) d \zz $, which is often intractable. However, in practice, for most $\zz$, $p_{\btheta}(\xx|\zz) \approx 0$, namely these $\zz$ provide minimal contribution in computing $p_{\boldsymbol{\theta}}(\xx)$. For this reason, the main idea of VAE is to sample values of $\zz$ that are likely to have produced $\xx$, and compute $p_{\boldsymbol{\theta}}(\xx)$ just from those. However, for doing that, we need to compute the posterior distribution $p_{\boldsymbol{\phi}}(\zz|\xx)$, which is also intractable. For this reason, VAEs rely on variational inference, which allows the approximation of the intractable posterior via $q_{\boldsymbol{\phi}}(\zz|\xx)$, known as recognition model in VAE parlance. To make the approximation feasible, $q_{\boldsymbol{\phi}}$ is assumed to follow a specific family of parametric distributions, usually a Gaussian distribution with 0 mean and unitary variance. The closeness between $q_{\boldsymbol{\phi}}(\zz|\xx)$ and the true posterior distribution $p_{\boldsymbol{\theta}}(\zz|\xx)$ is ensured by the minimization of the Kullback-Liebler divergence ($\KL$), which can be written as:
\begin{multline}
\label{eq:kl}
\KL(q_{\bphi}(\zz | \xx) \| p_{\btheta}(\zz | \xx))= \\
\mathbb{E}_{q_{\bphi}(\zz | \xx)}\left[\log q_{\bphi}(\zz | \xx) - \log p_{\btheta}(\xx, \zz)\right] + \log p_{\btheta}(\xx).
\end{multline}

After some rearrangements of Equation~\eqref{eq:kl}, it is possible to obtain to the so-called Evidence Lower BOund (ELBO)~\cite{kingma2014autoencoding}, which defines the objective function that a VAE tries to maximize:
\begin{equation}\label{eq:elbo}
\begin{aligned}
\log p_{\btheta}(\xx) & \geq \mathbb{E}_{q_{\bphi}(\zz | \xx)}\left[\log p_{\btheta}(\xx | \zz)\right] - \KL (q_{\bphi}(\zz | \xx) \| p_{\btheta}(\zz)) \\
& = \mathcal{L}\left(\xx; \btheta, \bphi\right).
\end{aligned}
\end{equation}

This equation can be interpreted as a reconstruction error (first term), plus the so-called $\KL$ loss, which acts as a regularization term.

The autoencoder comes into play when $q_{\bphi}$ and $p_{\btheta}$ are parameterized by two (deep) neural networks, i.e., the encoder ($f_{\bphi}$) and the decoder ($g_{\btheta}$), respectively. The parameters of $f_{\bphi}$ and $g_{\btheta}$ are optimized using stochastic gradient ascent with the aid of the reparameterization trick~\cite{kingma2014autoencoding}, which enables the computation of the gradient w.r.t. $\bphi$. To this end, given an input vector $\xx$, the encoder network provides the parameters which define the probability distribution over the $\zz$ that are likely to produce $\xx$, i.e, $q_{\bphi}(\zz|\xx)$. This Gaussian distribution is defined by the two outputs of the encoder, namely the mean $\boldsymbol{\mu}_{\bphi}(\xx)$ and the diagonal co-variance matrix $\boldsymbol{\Sigma}_{\bphi}(\xx)$. The sampling over this distribution is performed via an additional input $\boldsymbol{\epsilon}$, which allows the reparameterization 
$\zz = \boldsymbol{\mu}_{\bphi}(\xx) + \boldsymbol{\epsilon} \odot \boldsymbol{\Sigma}_{\bphi}(\xx)$, where $\odot$ is the Hadamard product.

\subsection{Auto-encoding Binary Classifiers}
Auto-encoding Binary Classifiers (ABC) \cite{yamanaka2019autoencoding} are supervised anomaly detectors based on the vanilla autoencoder. The authors' main contribution is the proposal of a hybrid approach in anomaly detection. Instead of using a purely supervised or unsupervised method, ABC exploits a binary classifier built in an autoencoder.
On the one hand, purely supervised methods have been shown to have low resilience to unknown anomalies; on the other hand, purely unsupervised methods are not able to distinguish negative from positive data, limiting their use in anomaly detection tasks. Using a hybrid approach makes it is possible to merge the advantages of the two methods while addressing their limitations. This integration has shown to be effective for performing anomaly detection.

The architecture of ABC is based on an autoencoder. We indicate with $f_{\bphi}$ the encoder and with $g_{\btheta}$ the decoder, parameterized by $\bphi$ and $\btheta$, respectively. ABC assumes that the data set $\XX = \{(\xx^{(i)}, y^{(i)})\}_{i=1}^N$ is subdivided in two sets: a set of ``normal'' data ($y=1$), and a set of anomalous data ($y=0$). The objective of ABC is to discriminate between normal data and anomalies. To achieve this, ABC uses the supervision given by $y$ to dynamically change the objective of the training depending on the class of the example. The main idea is to penalize the reconstruction of the anomalies, while minimizing the reconstruction error on normal data. Then, after the training, the model can be used to discriminate between anomalies and normal data by using the reconstruction error as a metric to detect anomalies.

The ABC loss function is defined as follows, where the reconstruction term for vector $\xx^{(i)}$ is defined as $\mathcal{L}_{AE}(\xx^{(i)}) = \|\xx^{(i)} - g_{\btheta}(f_{\bphi}(\xx^{(i)}))\|$, and $\|\cdot\|$ is an arbitrary distance function. The $\ell_2$-norm was used in the paper.
\begin{multline}\label{eq:abc}
-\log p_\theta(y^{(i)}|\xx^{(i)}) = \\
y^{(i)} \mathcal{L}_{\text{AE}} (\xx^{(i)}) - (1-y^{(i)}) \log (1-e^{-\mathcal{L}_{\text{AE}}(\xx^{(i)})})
\end{multline}

The first term has to be intended as a reconstruction error, while the second term can be interpreted as an anomaly reconstruction penalty. Given this objective, ABC will learn how to effectively reconstruct desired data while returning flawed reconstructions for anomalies.

In this paper, we show how this idea of penalizing anomalies can be used to avoid the generation of unwanted data on variational autoencoders.




\section{Proposed method}\label{sec:method}

This section introduces a novel problem formulation for generative methods and presents our V-ABC model.

\subsection{Problem formulation}

\begin{figure}[h!]
    \centering
    \includegraphics[width=0.35\textwidth]{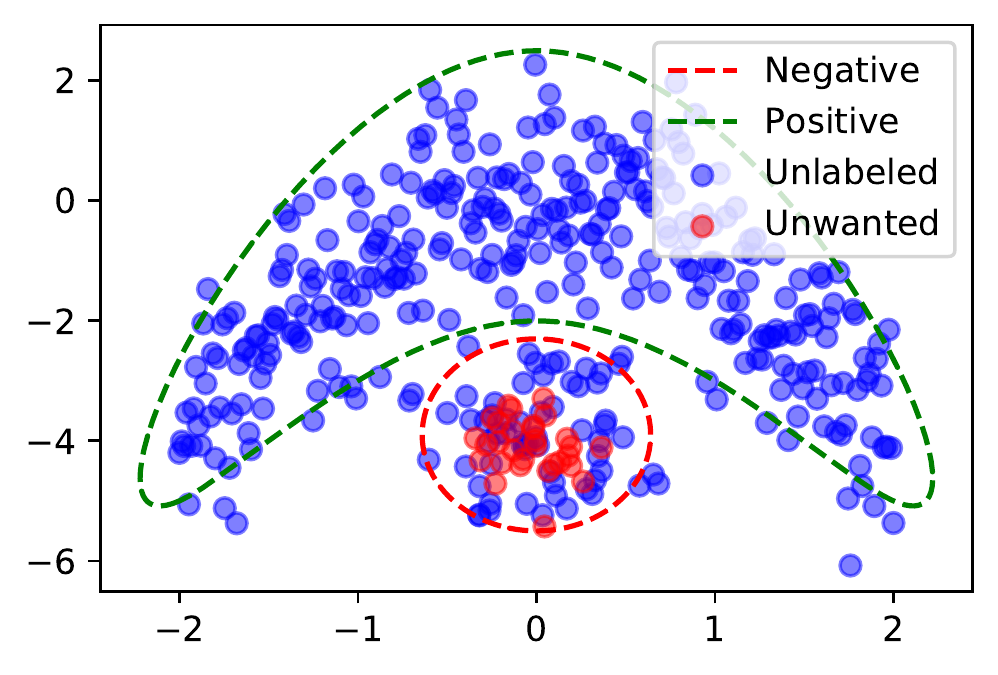}
    \caption{Possible instance of a \setting{} setting.}
\label{fig:prob_setting}
\end{figure}

We describe a novel \setting{} setting for generative methods. Let us assume that we have an unlabeled data set $\XX = \{\xx^{(i)}\}_{i=1}^N$, where $\xx^{(i)}$ is a vector of features describing the $i_{th}$ example. Then, suppose that our data set contains some examples that might be anomalous or undesired, for example, explicit songs or scary faces. We refer to these examples as unwanted data, in the sense the our model should avoid modeling and generating them. Figure~\ref{fig:prob_setting} depicts a possible instance of the \setting{} setting. The unlabeled examples are depicted with blue circles, while unwanted data is depicted by using red circles. Unwanted data may overlap with unlabeled data, which is what makes this setting challenging. For this reason, we define a \emph{negative concept}, namely an area in the (input) space in which unlabeled data lives together with unwanted data, and a \emph{positive concept}, namely an area of the space in which we have only evidence of unlabeled data. In the figure, the negative concept is surrounded by a red dashed line, while the positive concept is surrounded by a green dashed line. In general, these two concepts are defined by two different probability distributions. We denote with $p(\xx|+)$ the probability distribution which models the data set examples in the positive concept, and with $p(\xx|-)$ the probability distribution which models the data set examples in the negative concept. The objective of our approach is to learn $p(\xx|+)$ and use it for generating desired (i.e., not anomalous) data.

In this paper, we use a label $y$ to differentiate between unlabeled ($y=1$) and unwanted data ($y=0$). This leads us to a labeled data set, denoted with $\XX = \{(\xx^{(i)}, y^{(i)})\}_{i=1}^{N}$.

\subsection{Variational ABC}

Variational ABC (V-ABC) is an extension of ABC based on Variational Autoencoders, which enables generative modeling capabilities on the original ABC model. In particular, we show that V-ABC can be successfully used to avoid the generation of unwanted data.


Some practical observations drive the extension of ABC from an autoencoder-based architecture to a VAE-based one:
\begin{enumerate}
\item the ABC model is, at its heart, a classical autoencoder. It simply uses a label $y$ to define a different training objective for desired and anomalous data;
\item the conversion from autoencoders to VAEs occurs by first estimating the reconstruction error via a sampled latent vector $\zz$ and then adding a $\KL$-divergence regularization term.
\end{enumerate}

Given these observations, we define the loss function of a V-ABC model for a given input $(\xx^{(i)}, y^{(i)})$ as:
\begin{multline*}
\mathcal{L}_{\text{V-ABC}} = \KL(q_{\bphi}(\zz|\xx^{(i)})||p_{\btheta}(\zz))  \\
+ y^{(i)}\mathcal{L}_{\text{VAE}}(\xx^{(i)}) - (1-y^{(i)})\log(1-e^{-\gamma \mathcal{L}_{\text{VAE}}(\xx^{(i)})})
\end{multline*}

The first term is the classical $\KL$ loss of variational autoencoders, which forces the distribution $q_{\bphi}(\zz|\xx^{(i)})$ returned by the encoder to be similar to the prior $p_{\btheta}(\zz)$, while the second term can be interpreted as a reconstruction error which changes based on the value of $y^{(i)}$. Notice the similarities between this latter term and Equation~\eqref{eq:abc}. In particular, instead of $\mathcal{L}_{\text{AE}}$ we have $\mathcal{L}_{\text{VAE}}$, which is defined as follows.

\begin{equation*}
\mathcal{L}_{\text{VAE}}(\xx^{(i)}) = \frac{1}{2 \sigma^2}||\xx^{(i)}-g_{\btheta}(f_{\bphi}(\xx^{(i)}))||^2 
\end{equation*}

In this equation, $f_{\bphi}$ and $g_{\btheta}$ are the encoder and decoder of the V-ABC model, parameterized by $\bphi$ and $\btheta$, respectively. Then, $\sigma^{2}$ is a hyper-parameter deriving from the instantiation of the ELBO in Equation~\eqref{eq:elbo} for a Gaussian decoder  \cite{tutorialvae}, i.e., $\pxz \sim \mathcal{N}(g_{\btheta}(\zz), \sigma^2 * \mathbf{I})$.

Another critical difference between Equation~\eqref{eq:abc} and $\mathcal{L}_{\text{V-ABC}}$ is the presence of hyper-parameter $\gamma$ in the anomaly reconstruction penalty. The purpose of $\gamma$ is to calibrate the weight of unwanted examples ($y^{(i)}=0$) for V-ABC. As $\gamma$ increases, the model becomes more similar to a classical VAE. In particular, with $\gamma \rightarrow \infty$ V-ABC becomes a VAE modelling only positive samples ($y^{(i)}=1$). In this latter case, V-ABC will not enforce any reconstruction penalty for unwanted data but the $\KL$ term will constraint the latent distributions to be similar to the prior.

An advantage of this formulation is the ability to accurately adjust the distance we want to have from the \emph{negative concept} distribution $p(\xx|-)$. Higher values of $\gamma$ lead V-ABC to generate samples closer to unwanted data, while lower values allow generating samples further from unwanted data.

\section{Experiments}\label{sec:experiments}

In this section, we present the experiments performed with V-ABC. In particular, we conducted extensive trials on different toy and real-world data sets, seeking to answer the following research questions:

\begin{itemize}
    \item[\textbf{Q1}] Is V-ABC able to avoid the generation of unwanted data?
    \item[\textbf{Q2}] How do the different components of the model (e.g., encoder, decoder, labels) work together to avoid generating unwanted data? Does the latent space of V-ABC show any relevant property (e.g., continuity, completeness)?
    \item[\textbf{Q3}] Are the results consistent across different initial conditions (e.g., seeds, data set sizes, hyper-parameters choice)?
\end{itemize}

We made our source code\footnote{\url{tinyurl.com/2uh5xjzz}} available for reproducibility purposes. 


\subsection{Data sets}
We trained V-ABC on different toy and real-world data sets, summarized in Table~\ref{tab:datasets}. In the table, the number of examples in the positive ($\oplus$) and negative ($\ominus$) concepts are presented, as well as the number of unlabeled and unwanted examples, following the nomenclature proposed in Section~\ref{sec:method}. Specifically, $p$ is the proportion of negative examples considered as unwanted data during the training of V-ABC, denoted with $y=0$ in our problem formulation. The remaining negative examples are put in the unlabeled fold, together with the positive examples. As a reminder, the unlabeled examples are denoted with $y=1$. Our method is then trained on these two latter folds, namely the unlabeled and unwanted folds. The \textit{class} column in the table indicates which class from the data set has been chosen as negative concept. In the following, we refer to negative data as the examples in the negative concept, and to positive data as the examples in the positive concept.

\begin{table}[h]
\caption{Data sets information} 
\label{tab:datasets}

\centering
\begin{tabular}{ l c c c c c c }
\hline
Data set & class & p & \#$\oplus$ & \#$\ominus$ & \#unlabeled & \#unwanted  \\
 \hline
 \texttt{Moons1} & 0 & 1 & 5000 & 5000 & 5000 & 5000 \\ 
 \texttt{Moons2} & 0 & 0.8 & 5000 & 5000 & 5983 & 4017 \\ 
 \texttt{MNIST} & 1 & 0.1 & 53258 & 6742 & 59328 & 672 \\ 
 \texttt{MNIST} & 1 & 0.2 & 53258 & 6742 & 58686 & 1314 \\ 
 \texttt{MNIST} & 1 & 0.3 & 53258 & 6742 & 57986 & 2014 \\ 
 \texttt{MNIST} & 7 & 0.1 & 53735 & 6265 & 59375 & 625 \\ 
 \texttt{MNIST} & 7 & 0.2 & 53735 & 6265 & 58777 & 1223 \\ 
 \texttt{MNIST} & 7 & 0.3 & 53735 & 6265 & 58125 & 1875 \\ 
 \texttt{MNIST} & 8 & 0.1 & 54149 & 5851 & 59414 & 586 \\ 
 \texttt{MNIST} & 8 & 0.2 & 54149 & 5851 & 58858 & 1142 \\ 
 \texttt{MNIST} & 8 & 0.3 & 54149 & 5851 & 58276 & 1724 \\ 
 \hline
\end{tabular} 
\vspace{12pt}
\end{table}

\paragraph{Moons1} this synthetic data set is composed by two interleaving half circles, also referred as \textit{moons} by the clustering research community. The lower moon contains only positive data, while the upper moon contains negative data. This data set presents a minimal, but existing, overlapping between the two distributions (i.e., lower and upper moons). \texttt{Moons1} constitutes an \textit{easy} setting, in the sense that the unlabeled and unwanted folds contain exactly the positive and negative data, respectively.

\paragraph{Moons2} the previous toy data set is extended to an harder formulation, where the unlabeled fold contains a portion (20\%) of negative data, while the remainder (80\%) is moved on the unwanted data fold.

\paragraph{MNIST} to show the versatility of V-ABC on real-world scenarios, we employ a simple computer vision data set consisting of images of handwritten digits. The data set is already subdivided into training and test sets. The training examples related to one digit have been considered as negative data, while all the others as positive data. Then, 10\% of the negative data has been selected as unwanted data for V-ABC, while the remaining 90\% composes the unlabeled fold together with the positive examples. In our experiments, we tried different values for $p$ (i.e., $0.1,0.2,0.3$) and selected `1', `7', and `8' as negative data in three different trials.

\subsection{Model architecture}
Depending on the data set, we used different architectures for V-ABC. 
\paragraph{Moons} We used a simple Multi-Layer Perceptron (MLP) as encoder and decoder, with two hidden layers with 20 units each. The latent space dimension has been set to 1. \texttt{ReLU} has been used as activation function for the hidden layers. For the output of the encoder, we used a linear activation. The same is done for the output of the decoder. The final output has been chosen to be linear since we interpret it as the mean of the Gaussian distribution modelled by the probabilistic decoder. A sample can be then visualized by sampling from the aforementioned distribution, using the predicted mean and by using 1 as variance, as we will see in our qualitative results. During training, we fixed hyper-parameter $\sigma^2 = 1$.

\paragraph{MNIST} We increased the number of hidden units to 300 and 100 for the outer and inner layers, respectively. In this case, a sigmoid function has been applied to the output of the network to compute pixel values between 0. and 1. We fixed $\sigma^2 = 2.5$.

\subsection{Training procedure}
During training, we balanced the unlabeled and unwanted data in the mini-batches to mitigate the effect of unbalanced data sets, by ensuring to sample an equal number of unwanted and unlabeled examples for each mini-batch. We used a sigmoidal annealing schedule \cite{bowman2016generating} for hyper-parameter $\gamma$, described as follows.

\begin{itemize}
    \item at the beginning of the training, we set $\gamma$ to a relatively high value (e.g., 4). This helps the model in accurately learn the positive distribution without penalizing it for reproducing negative samples in the early stages of training;
    \item during training, we decrement $\gamma$ by using a sigmoidal scheduling, forcing the model to progressively consider negative examples and learn how to not generate them;
    \item $\gamma$ is reduced until it reaches its desired pre-set final value (e.g., 1).
\end{itemize}

For the annealing of the $\KL$ term, we followed the procedure described by \cite{bowman2016generating}. The annealing for the $\KL$ term and $\gamma$ has been performed only on the epochs of training. We refer to this number of epochs as $E'$.

In addition, for \texttt{MNIST} we used early stopping as a regularization technique, using the test data set to compute a validation score. The test loss function has been used as validation metric. It is important to highlight that the test set has been exclusively used for validation purposes.

For each experiment, we used batch size equal to 80, and \texttt{Adam} as an optimizer, with parameters set to default values. For \texttt{Moons}, we trained V-ABC for 30 epochs with $E'=10$ and $\gamma=3$, while for \texttt{MNIST} we used $E'=5$ and $\gamma=0.05$.

\subsection{Evaluation procedure}
We evaluated the results on the \texttt{Moons} data sets in a qualitative manner, by plotting the sampled data over the original data. For the \texttt{MNIST} data set, we used a quantitative evaluation. In particular, a Convolutional Neural Network (CNN) classifier has been trained on the original positive and negative labeled data. Then, this classifier is used to measure the correctness of the generated samples after the training of V-ABC. For each \texttt{MNIST} data set in Table~\ref{tab:datasets} we repeated the training procedure using 3 different random seeds to investigate possible sensitivity to initial conditions.

We compared V-ABC to a vanilla VAE. We trained the VAE on the unlabeled fold of the data sets by using the same architecture as for V-ABC.

\section{Results}\label{sec:results}

This section presents the results of the experiments conducted, subdividing them by research question and data set.

\subsection{Negative concept avoidance (\textbf{Q1})}
In this section, we analyze the ability of V-ABC to avoid generating negative data by evaluating the output of our model in both a qualitative and quantitative way.

\begin{figure*}[!tbp]
  \centering
  \begin{minipage}[h]{0.485\textwidth}
    \includegraphics[width=\textwidth]{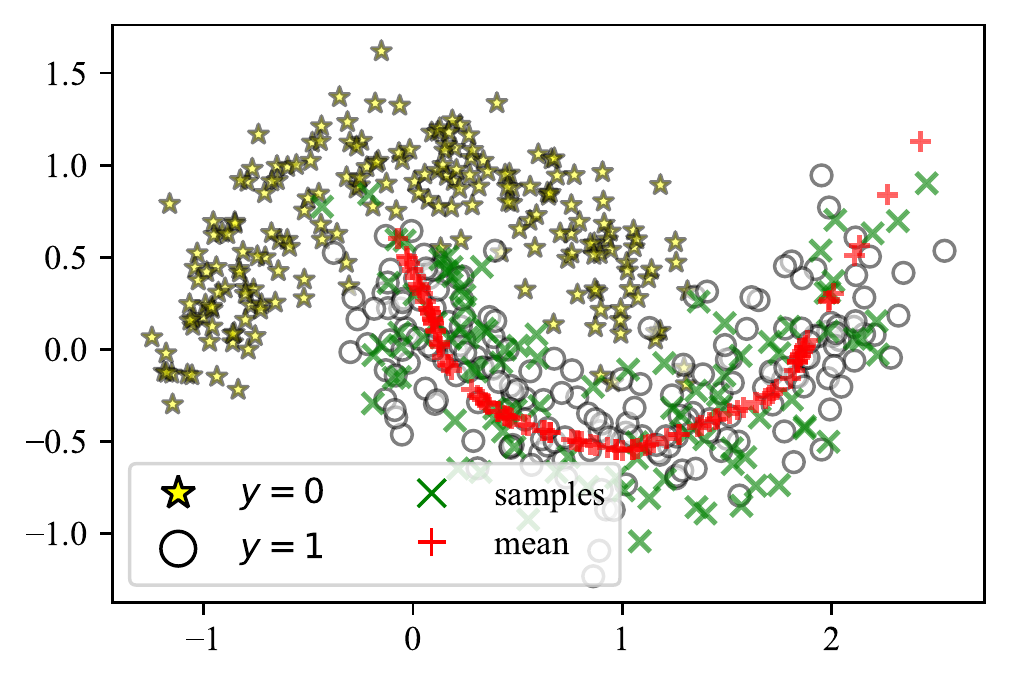}
  \end{minipage}
  \hfill
  \begin{minipage}[h]{0.485\textwidth}
    \includegraphics[width=\textwidth]{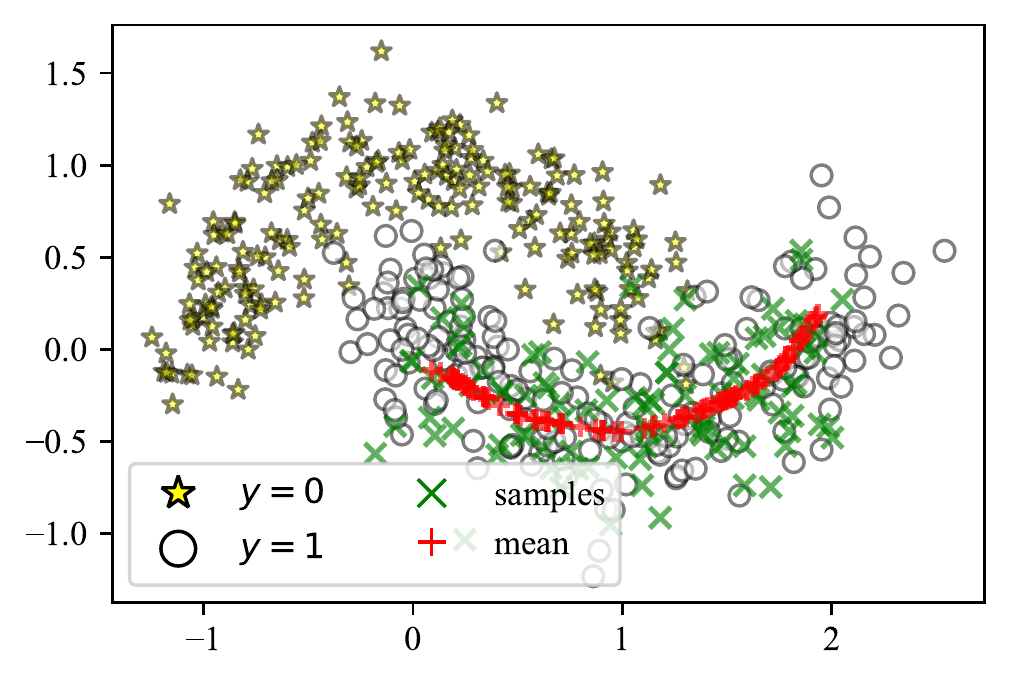}
  \end{minipage}
  \caption{VAE (left) and V-ABC (right) model samples; \texttt{Moons1} data set. Green crosses represent sampled data. Red plus signs represent the mean of the distributions used for sampling. $y=0$ represents unwanted data, $y=1$ represents unlabeled data. We display only 100 data points for each class to avoid clutter.}
\label{fig:easy}

\end{figure*}

\begin{figure*}[!tbp]
  \centering
  \begin{minipage}[h]{0.485\textwidth}
    \includegraphics[width=\textwidth]{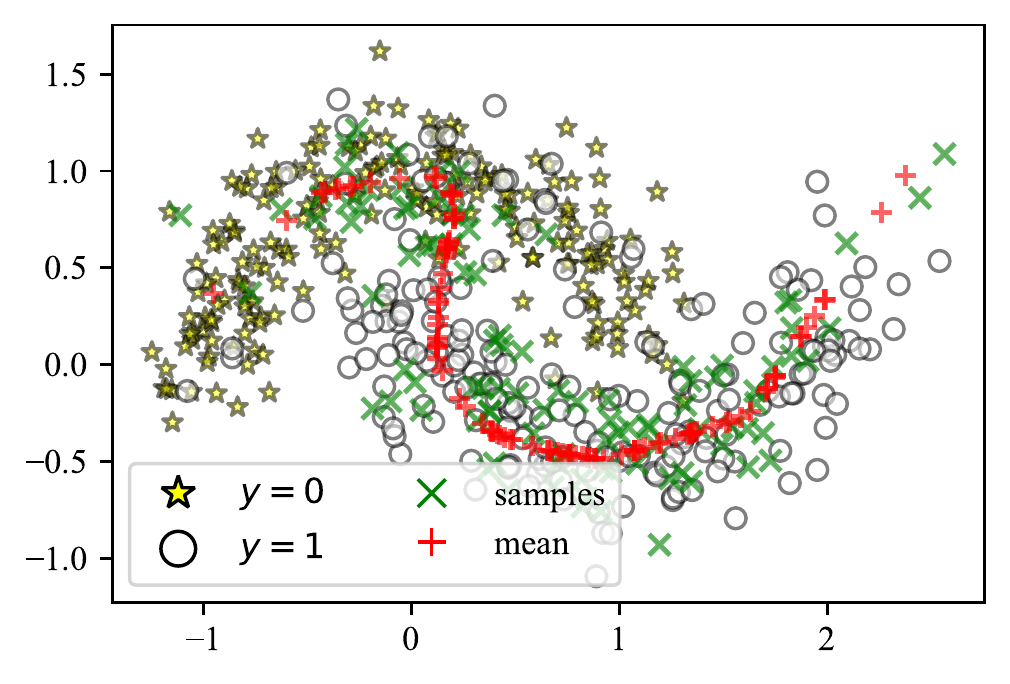}
  \end{minipage}
  \hfill
  \begin{minipage}[h]{0.485\textwidth}
    \includegraphics[width=\textwidth]{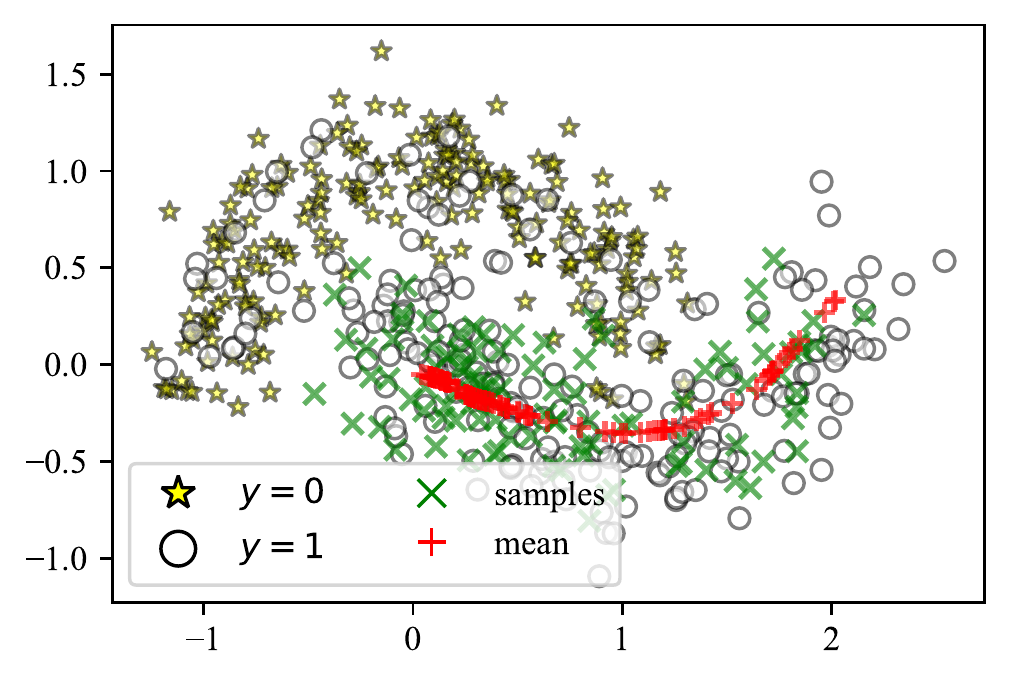}
  \end{minipage}
  \caption{VAE (left) and V-ABC (right) model samples; \texttt{Moons2} data set. Green crosses represent sampled data. Red plus signs represent the mean of the distributions used for sampling. $y=0$ represents unwanted data, $y=1$ represents unlabeled data. We display only 100 data points for each class to avoid clutter.}    \label{fig:hard}

\end{figure*}

\paragraph{Moons}
As shown in Figure~\ref{fig:easy} (\texttt{Moons1}), a VAE can faithfully reproduce the selected distribution, but it occasionally creates data that is similar to the negative concept. Additionally, it can possibly create outliers. \method{} increases the distance of the generated samples from the negative distribution by avoiding generating data near to the upper left of the lower moon, since it is surrounded by negative data. In addition, it also reduces the generation of outliers on the right side of the lower moon. Despite unsupervised models are feasible approaches for sufficiently separated data sets, the \setting{} setting is unfeasible to solve for a classic VAE. In fact, as shown in Figure~\ref{fig:hard} (\texttt{Moons2}), the VAE spans the whole unlabeled distribution. The V-ABC, instead, has a preferable behavior, since it learns how to avoid the upper moon and similar unlabeled data in the process. We observe that the probability mass modeled by V-ABC is simply moved to a safer place, increasing the frequency of generating data points on the left. This is noteworthy, since one may be interested in generating data that is similar as much as possible to the unlabeled data distribution, without actually creating new negative data.

\begin{figure}[h!]
    \centering
    \includegraphics[width=0.35\textwidth]{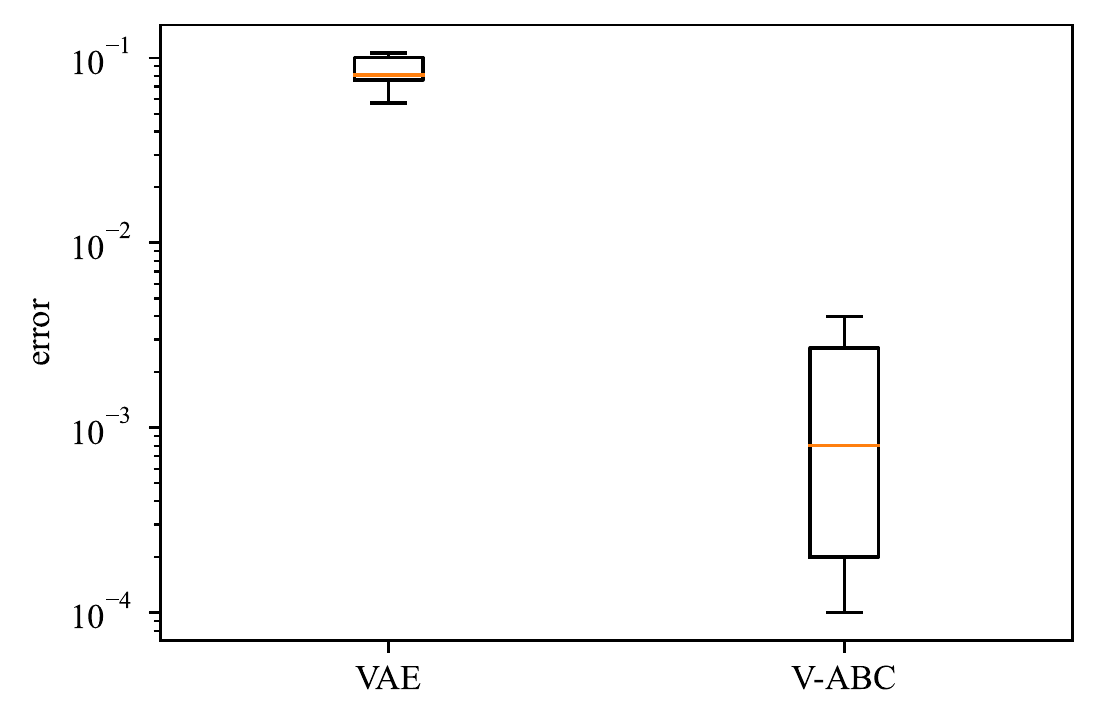}
    \caption{Negative generation error for \texttt{MNIST} data set ($p=0.2$); VAE vs. V-ABC.}
\label{fig:gen-error-mnist}
\end{figure}

\paragraph{MNIST}
The performances on \texttt{MNIST} have been quantitatively evaluated by using a CNN classifier to 
assess the frequency of unwanted data generated by V-ABC. We denote this metric as \textit{negative generation error}.
The classifier used shows an average negative generation error equal to $0.14\% \pm 0.0016$ for V-ABC, as opposed to $8.31\% \pm 0.0181$ for VAE, as reported in Figure~\ref{fig:gen-error-mnist}.

\begin{figure*}[!tbp]
  \centering
  \begin{minipage}[h]{0.485\textwidth}
    \includegraphics[width=\textwidth]{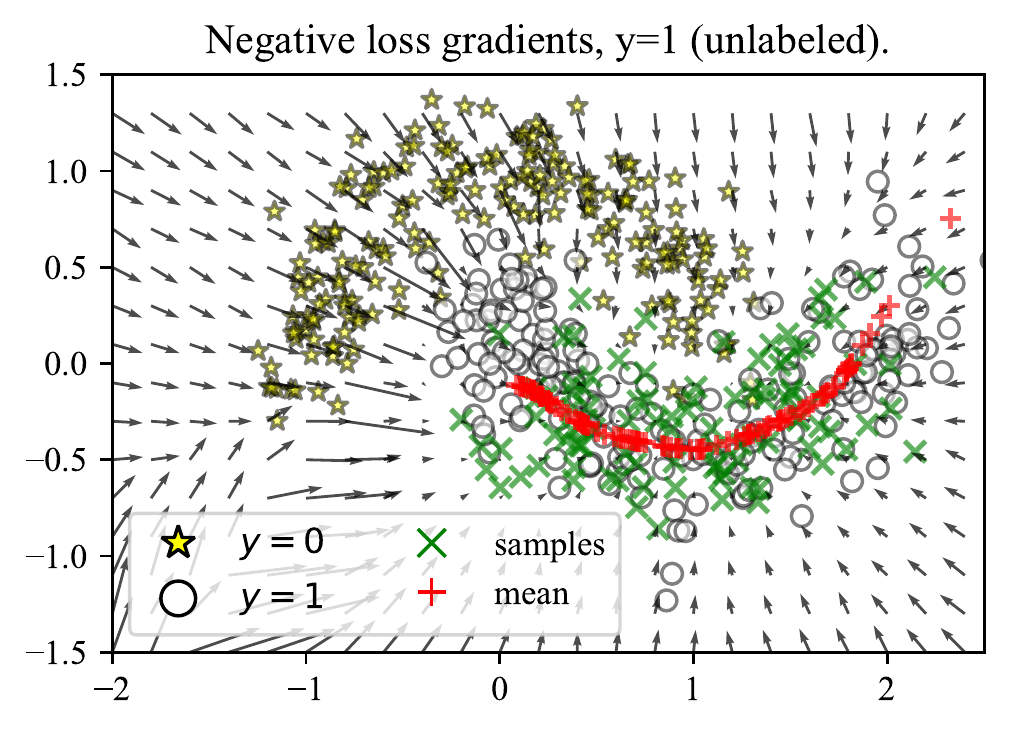}
  \end{minipage}
  \hfill
  \begin{minipage}[h]{0.485\textwidth}
    \includegraphics[width=\textwidth]{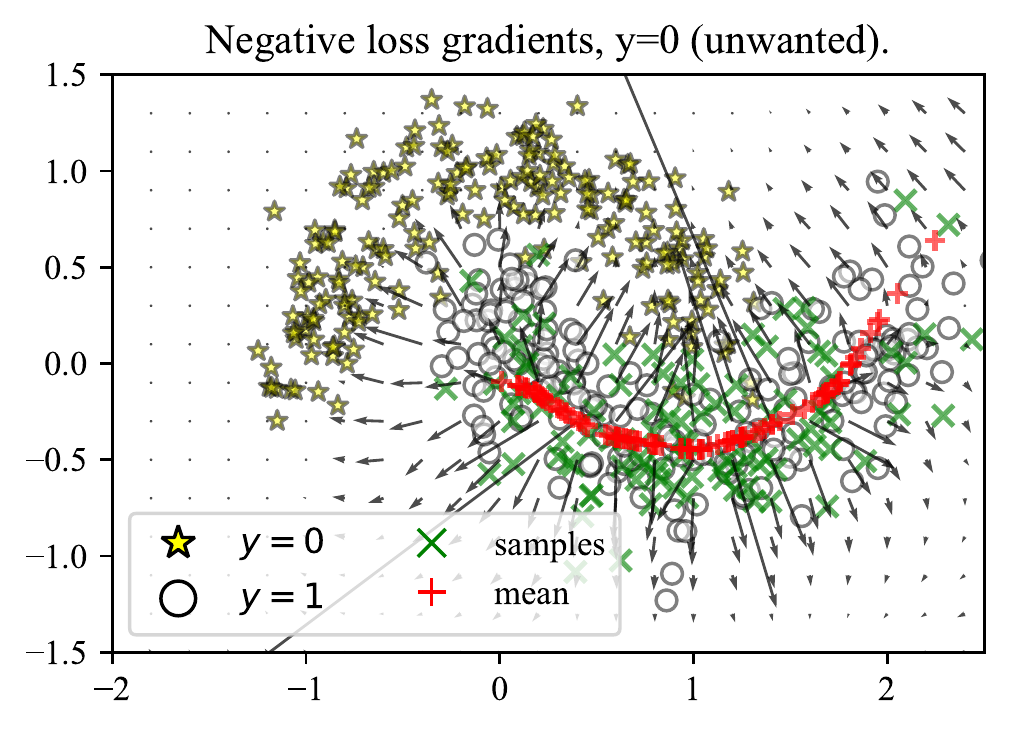}
  \end{minipage}
  \caption{Vector field of the gradient of $\mathcal{L}_{\text{V-ABC}}$ w.r.t. the input grid, given unlabeled input $y=1$ (left), and unwanted input $y=0$ (right). We display only 100 data points for each class to avoid clutter.}
  \label{fig:gradients}
\end{figure*}


		

\subsection{Model analysis (\textbf{Q2})}
We exploited different visualization techniques to better understand the model components. 

\paragraph{Moons}
A recent paper on GAN \cite{thanhtung2020catastrophic} shows an effective way to visualize better the impact of a loss function on a 2D data set. We applied this method to a VAE using a simplified formulation. In particular, we computed the gradient $-\frac{\partial{\mathcal{L}_{\text{V-ABC}}(\xx)}}{\partial \xx}$ and found the direction of the steepest descent of the loss function. By computing the gradient of the loss function w.r.t. an input grid, we created a \textit{vector field}, depicted in Figure~\ref{fig:gradients}. To compute the gradient, we skipped the stochastic step in the model by choosing $\zz = \boldsymbol{\mu}_{\bphi}(\xx)$. Avoiding sampling makes the gradients behave as they would do on average.

The computation of the gradient allows to give an interpretation of how the model expects the data points to be positioned in the space. High gradients show that the model expects a data point to be positioned in another location, while low gradients show that the model is confident that a point is correctly positioned. As we can see in the figure, the gradient pushes unlabeled data to be positioned near the lower moon, while unwanted data has high gradients only for those data points that the model is extremely confident they belong to the positive distribution.








\paragraph{MNIST} 

\begin{figure*}[!h]
  \begin{minipage}[h]{0.5\textwidth}
  \centering
    \includegraphics[width=0.70\textwidth]{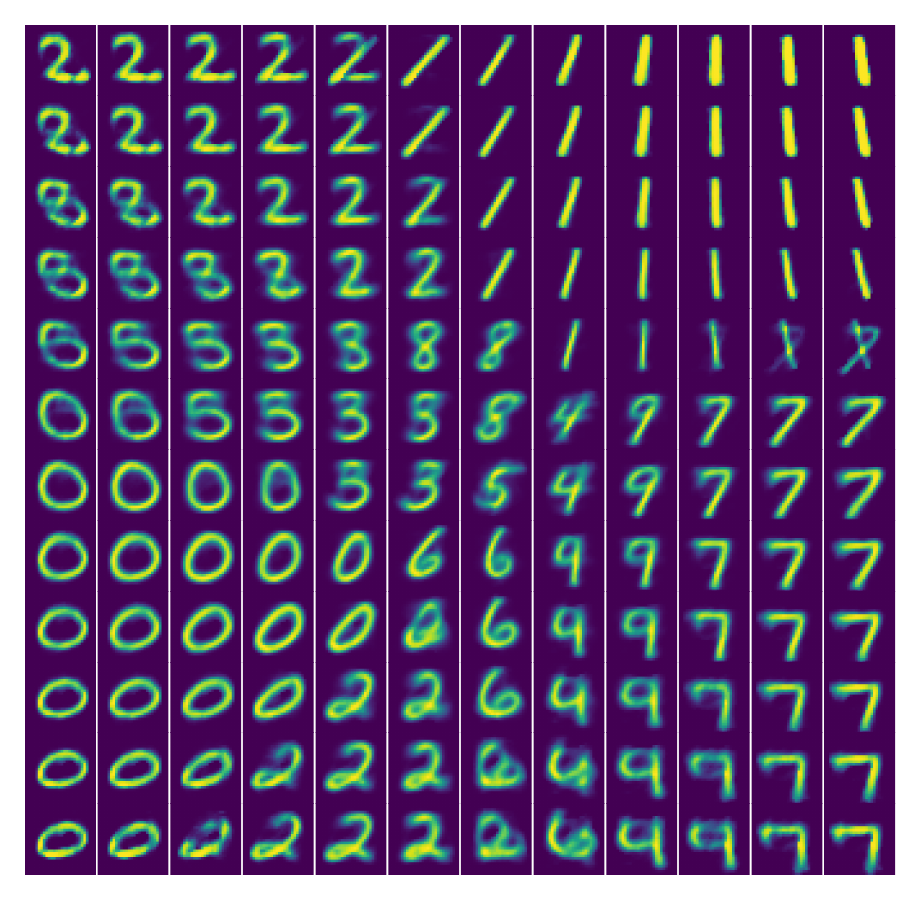}
  \end{minipage}
  \hfill
  \begin{minipage}[h]{0.5\textwidth}
  \centering
    \includegraphics[width=0.70\textwidth]{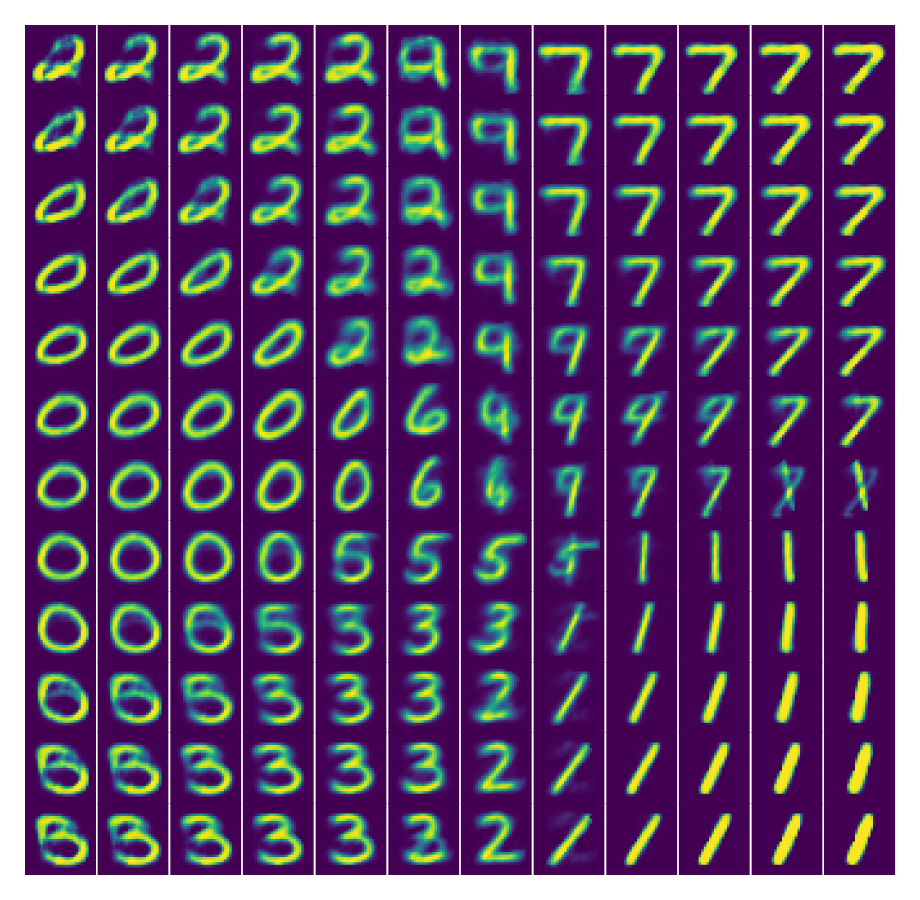}

  \end{minipage}
  \caption{Manifold: VAE (left) vs. V-ABC (right); \texttt{MNIST} data set ($p=0.2$). The unwanted digit `8' is removed by V-ABC.} 
  \label{fig:manifold}
\end{figure*}

By using two latent dimensions, we can easily represent the latent space and create a two-dimensional manifold for \texttt{MNIST}. To plot the manifold, we selected different values from a grid in the interval $[-3,3]$ for each dimension of $\zz$, then, we applied the decoder to each $\zz$ value. By using this method, it is possible to inspect the behavior of the decoder, giving an intuitive way to reason on $\pxz$. For example, we may assess the area associated with each digit, e.g. evaluating whether a digit is more frequent than another, in a visual way. We can also get a general idea of the sample quality.

Figure~\ref{fig:manifold} shows how the V-ABC latent space exhibits a structure which is similar to the classic VAE counterpart. By traversing it along any direction, digits slowly transform in shape, stroke, or angle. A classic VAE can generate only the digits supplied, making it useful and effective for fully labeled data sets. In the \setting{} setting, this model has a clear disadvantage: the model cannot discern unwanted digits, limiting itself to imitate the input distribution that contains all the digits. The V-ABC manifold displays an important difference: the unwanted digit `8' is not generated. 

\begin{figure}[!h]
  \centering
  \begin{minipage}[h]{0.4\textwidth}
    \includegraphics[width=\textwidth]{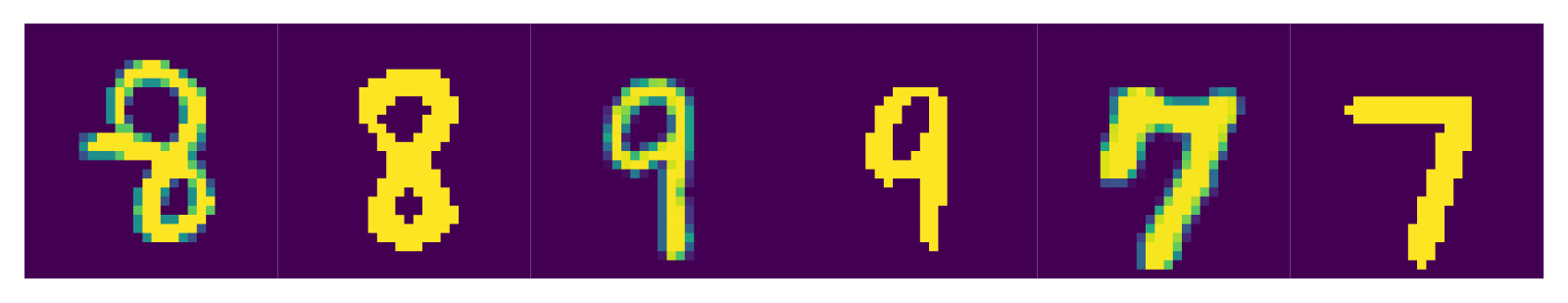}
\label{fig:recon-vae}  
  \end{minipage}
  \hfill
  \begin{minipage}[h]{0.4\textwidth}
    \includegraphics[width=\textwidth]{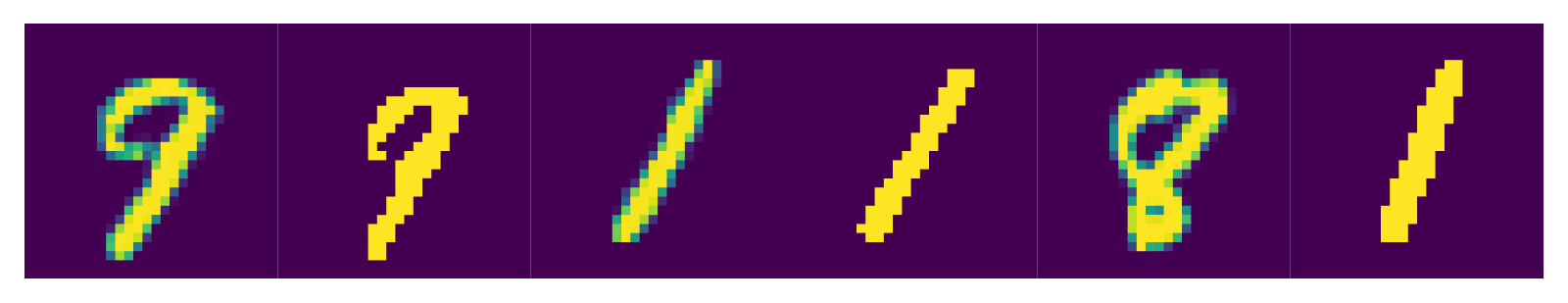}

  \end{minipage}
  \caption{Reconstruction: VAE (upper) vs. V-ABC (lower); \texttt{MNIST} data set ($p=0.2$). First, third, and fifth columns: original data. Second, fourth, and sixth columns: reconstructed data (threshold filter applied). The unwanted digit `8' is reconstructed to the same digit only for the VAE.}
  \label{fig:recon}  

\end{figure}

Another way to assess an autoencoding model is to visualize how data is reconstructed. An autoencoder is expected to reconstruct data given as input, even if occluded or corrupted. Figure~\ref{fig:recon} shows, in practice, how a VAE and V-ABC reconstruct the inputs. For the VAE, the digits are reconstructed correctly, except whenever a digit is misread; in this case, a different, plausible digit is reconstructed. The V-ABC can reconstruct input data just as well, but with a fundamental difference: it reconstructs unwanted digits as different, plausible digits. The positive data is recostructed, instead, correctly, similarly to the VAE. 

\begin{figure}[h!]
    \centering
    \includegraphics[width=0.45\textwidth]{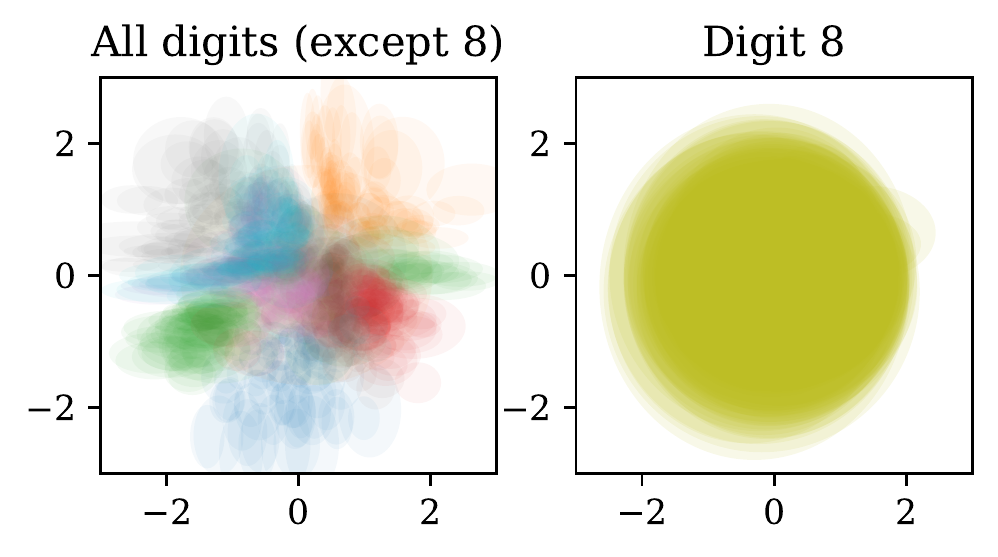}
    \caption[Encoder, V-ABC.]{$\boldsymbol{\mu}_{\bphi}(\xx)$ and $\boldsymbol{\Sigma}_{\bphi}(\xx)$ visualization, V-ABC; \texttt{MNIST} data set ($p=0.2$). Each color represents a different digit. Each ellipsis represents the area in $\boldsymbol{\mu}_{\bphi}(\xx) \pm 2\boldsymbol{\Sigma}_{\bphi}(\xx)$. `8' is the unwanted digit.}
\label{fig:encoder-vabc}
\end{figure}

In Figure~\ref{fig:encoder-vabc}, we visualize the encoder behavior by computing $\boldsymbol{\mu}_{\bphi}(\xx)$ and $\boldsymbol{\Sigma}_{\bphi}(\xx)$ on a training data subset. Hence, we give an intuitive way to reason on the $\qzx$ distribution. These values are visualized by creating an ellipse for each data point considered; its mean locates its center, and its variance gives the width and the height to the ellipse. The reparameterization step in a VAE-based model selects with high probability\footnote{The area of each ellipse shows $2 \sigma$, corresponding to about 95\% of data.} a $\zz$ inside this ellipse. The distribution $\qzx$ tries to be similar to the Gaussian prior $p_{\btheta}(\zz)$, having mean $\mathbf{0}$ and diagonal co-variance matrix $\mathbf{I}$. In the figure, it is possible to observe that different latent regions encode different digits, with some overlapping because of ambiguous data. Negative digits instead have a different behavior: in this case, the distribution $\qzx$ is much closer to the prior, as it covers the whole distribution. As long as the positive data covers well the whole latent space, the decoder cannot reconstruct negative data, and instead picks up a random $\zz$ vector, drawn by the uninformative prior.






\subsection{Sensitivity to initial conditions (\textbf{Q3})}

Finally, to test the limits of V-ABC, we trained different model instances to assess its robustness to different initial conditions (\textbf{Q3}).

\begin{figure}[!h]
  \centering
  \begin{minipage}[h]{0.22\textwidth}
    \includegraphics[width=\textwidth]{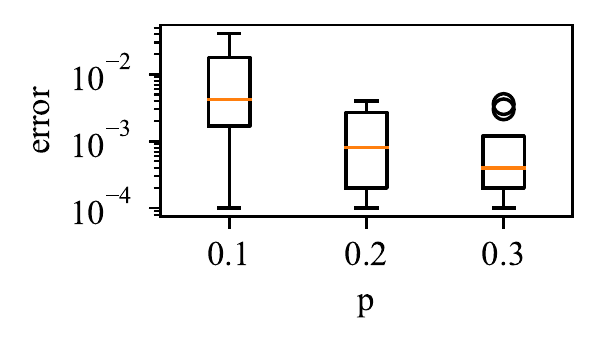}
  \end{minipage}
  \hfill
  \begin{minipage}[h]{0.22\textwidth}
    \includegraphics[width=\textwidth]{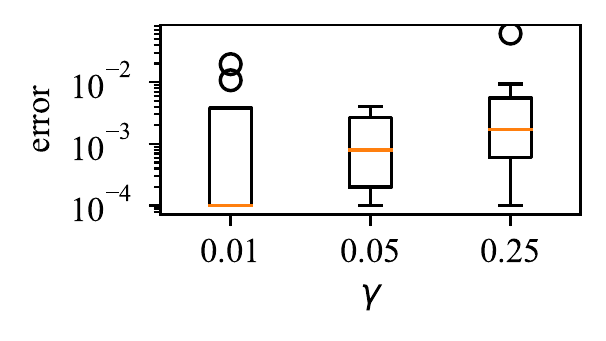}

  \end{minipage}
  \caption{Negative generation error for different unwanted data set sizes (left), and $\gamma$ values (right).} 
  \label{fig:boxplot}
\end{figure}

\paragraph{Moons} V-ABC is resilient to the different initial conditions tested, since it gives similar results by changing the random seed and $p$. By increasing $\gamma$, the model reduces its distance from the negative data distribution, while by decreasing it, the model increases this distance. A very low $\gamma$ leads to clear distortion, which changes the shape of the positive moon to further increase the distance from negative data.

\paragraph{MNIST}
We investigated how the performance varies using different sizes for the unwanted data set. The first plot in Figure~\ref{fig:boxplot} shows that the use of more unwanted samples leads to lower negative generation error, while a less number of unwanted samples leads to a much higher error and variance, showing that there is a clear trade-off between data set size and the model capacity to avoid generating unwanted data.

Finally, we show how the performance changes using different values for $\gamma$. As shown in the second plot in Figure~\ref{fig:boxplot}, a higher $\gamma$ increases the overall error. Instead, a lower $\gamma$ can reduce the average error. However, it is more sensitive to initial conditions, as shown by the higher scores for outliers. This behavior decreases the reliability of the model found.

\section{Conclusions}\label{sec:conclusions}

In this paper, we showed it is possible to control the generative capabilities of Variational Autoencoders. In particular, we repurposed the autoencoding-based ABC anomaly detector, extending it by regularizing its latent space. This novel model, named V-ABC, has proved to be effective in tasks dominated by uncertainty, as we showed in our experimentation in the \setting{} setting. We also contributed to the ABC model by proposing an elegant solution to weight negative samples differently using $\gamma$, and by applying annealing techniques for training the model. This resulted in better performances and training behavior for the V-ABC model. 
As shown by the experimental results, the value of $\gamma$ is critical to the overall sample quality. Hence, it is needed to select it by adopting a rigorous hyper-parameter selection process.


We believe that further exploration of the \setting{} setting is imperative. Future work will focus on experimenting with other state-of-the-art VAE-based anomaly detectors to illustrate their generative capabilities. 


\bibliographystyle{IEEEtran}
\bibliography{ijcnn22}

\end{document}